\documentclass[11pt,draftcls,onecolumn,journal]{IEEEtran}
\usepackage{cite}
\usepackage{booktabs}
\usepackage{multirow}
\usepackage{geometry}
\usepackage{adjustbox}
\usepackage{amsmath,amsfonts}
\usepackage{algorithmic}
\usepackage{algorithm}
\usepackage{array}
\usepackage[caption=false,font=normalsize,labelfont=sf,textfont=sf]{subfig}
\usepackage{textcomp}
\usepackage{tabularx}
\usepackage{stfloats}
\usepackage{listings}
\usepackage{url}
\usepackage{verbatim}
\usepackage{graphicx}
\usepackage{cite}
\usepackage{xcolor}
\usepackage{longtable}
\usepackage{geometry}
\usepackage{color}
\usepackage{colortbl}
\usepackage{hyperref}
\usepackage{booktabs}
\begin{document}

\title{From Correctness to Comprehension: AI Agents for Personalized Error Diagnosis in Education}

\author{Yi-Fan Zhang, Hang Li, Dingjie Song, Lichao Sun, Tianlong Xu, Qingsong Wen$\dagger$
\thanks{$\dagger$: corresponding author;}
\thanks{Yi-Fan Zhang is with the National Laboratory of Pattern Recognition, University of Chinese Academy of Sciences}
\thanks{Hang Li is with the Computer Science department from Michigan State University.}
\thanks{Dingjie Song is with the CUHK-Shenzhen Natural Language Processing group.}
\thanks{Lichao Sun is with the Computer Science and Engineering at Lehigh University.}
\thanks{Tianlong Xu and Qingsong Wen are with the Squirrel Ai Learning group.}

}


\maketitle

\section*{}
\label{sec:abstract}
Large Language Models (LLMs), such as GPT-4, have demonstrated impressive mathematical reasoning capabilities, achieving near-perfect performance on benchmarks like GSM8K. However, their application in personalized education remains limited due to an overemphasis on correctness over error diagnosis and feedback generation. Current models fail to provide meaningful insights into the causes of student mistakes, limiting their utility in educational contexts. To address these challenges, we present three key contributions. First, we introduce \textbf{MathCCS} (Mathematical Classification and Constructive Suggestions), a multi-modal benchmark designed for systematic error analysis and tailored feedback. MathCCS includes real-world problems, expert-annotated error categories, and longitudinal student data. Evaluations of state-of-the-art models, including \textit{Qwen2-VL}, \textit{LLaVA-OV}, \textit{Claude-3.5-Sonnet} and \textit{GPT-4o}, reveal that none achieved classification accuracy above 30\% or generated high-quality suggestions (average scores below 4/10), highlighting a significant gap from human-level performance. Second, we develop a sequential error analysis framework that leverages historical data to track trends and improve diagnostic precision. Finally, we propose a multi-agent collaborative framework that combines a Time Series Agent for historical analysis and an MLLM Agent for real-time refinement, enhancing error classification and feedback generation. Together, these contributions provide a robust platform for advancing personalized education, bridging the gap between current AI capabilities and the demands of real-world teaching.


\section{Introduction}
\begin{figure}
    \centering
    \includegraphics[width=0.9\linewidth]{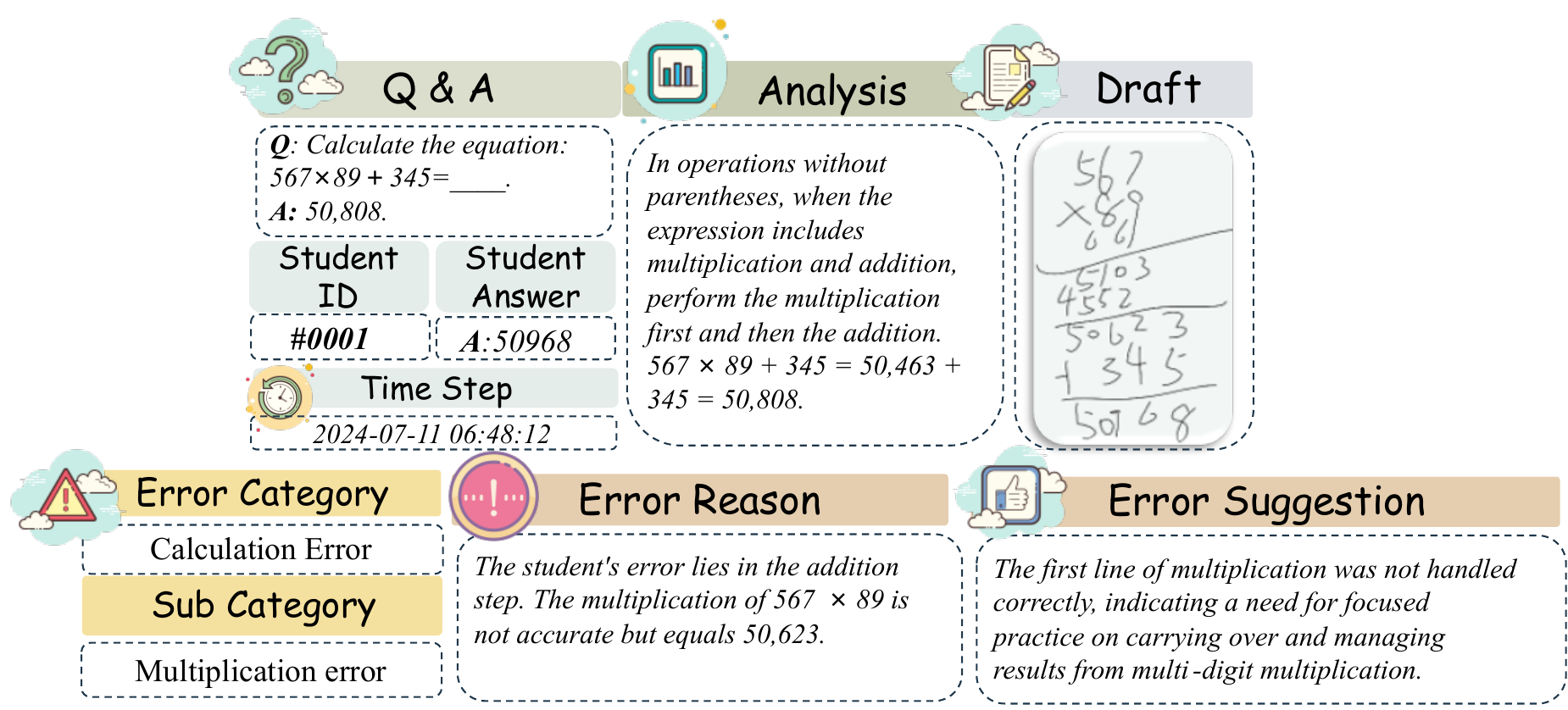}
    \caption{Each data sample in the dataset includes traditional question-answer pairs along with students' responses. Additionally, we provide students' drafts and a detailed analysis of the problems to furnish the model with more contextual information. Each student has multiple time-step data points, which support the construction of user profiles and enable the delivery of personalized recommendations. Finally, we include annotations from educational experts that identify the reasons for errors in the problems, along with relevant suggestions for improvement.}
    \label{fig:data}
\end{figure}
Large Language Models (LLMs), such as GPT-4~\cite{gpt4}, have achieved remarkable advancements in mathematical reasoning, particularly in solving mathematical word problems (MWPs)~\cite{roy2018mapping}. For instance, GPT-4 achieves a 97\% accuracy on the GSM8K dataset~\cite{zhou2023solving}, showcasing its ability to comprehend complex numerical contexts and execute multi-step reasoning. However, existing research primarily focuses on evaluating models based on the correctness of their answers and the consistency of their intermediate reasoning steps. While these metrics are important, they fail to address the critical challenge of diagnosing student errors and providing actionable, constructive feedback—an essential component of personalized education~\cite{prochazka2021integrating,gannot2023data,wang2024large,xu2024ai}. For educators, determining whether an answer is correct is often straightforward, as ground truth responses are typically predefined. The true challenge lies in understanding why students make mistakes and tailoring feedback to address these errors. Despite the advancements of LLMs and Multi-Modal Large Language Models (MLLMs), several key limitations hinder their application in personalized education:
\begin{itemize}
    \item \textbf{Overemphasis on binary correctness:} Current models primarily assess whether answers are correct or not, providing little insight into the reasoning behind mistakes. Such binary evaluations fall short in identifying nuanced error patterns and offering meaningful feedback.
    \item \textbf{Limited use of historical context:} Many educational tools fail to consider students' learning histories, overlooking the sequential and contextual nature of problem-solving. Historical performance data is critical for understanding learning patterns and constructing user profiles to improve feedback relevance~\cite{meier2015predicting,xu2017machine}.
    \item \textbf{Inadequate support for open-ended reasoning:} Existing evaluation systems often simplify assessments by fixing responses to predefined options~\cite{zhang2024mme} or short phrases~\cite{singh2019towards}. These methods fail to capture the complexity of open-ended problem-solving, where errors may arise from diverse reasoning pathways.
\end{itemize}

To address these challenges, we present several novel contributions aimed at advancing personalized education through systematic error analysis and tailored feedback. These contributions can be summarized as follows:

1. \textbf{The MathCCS Benchmark:}  
   We introduce \textbf{MathCCS} (Mathematical Classification and Constructive Suggestions), a multi-modal and multi-type error analysis benchmark that integrates real-world mathematical problems, student responses, and expert annotations. MathCCS includes:
   \begin{itemize}
       \item \textbf{Real-world problems and responses:} Mathematical tasks and applied scenarios, accompanied by students' drafts, responses, and detailed analyses.
       \item \textbf{Expert annotations:} Errors categorized into five major categories and forty subcategories, along with constructive suggestions tailored to each error type.
   \end{itemize}
   We evaluate MathCCS using several state-of-the-art models, including \textit{Qwen2-VL}, \textit{GPT-4o}, and \textit{Claude3.5-sonnet}. None of these models achieves a classification accuracy exceeding 30\%. Furthermore, the quality of suggestions generated by MLLMs is consistently poor, with average scores below 4 (on a scale of 10). These results highlight a significant gap between current MLLMs and human educators, underscoring the need for more robust and effective systems.

2. \textbf{Sequential Error Analysis Framework:}  
   Beyond MathCCS, we develop a sequential dataset to evaluate how historical data impacts models' performance in error diagnosis. This dataset enables: 1. Identification of trends in student performance over time; and 2. Extraction of historical patterns to diagnose errors more accurately.
   
   This sequential dataset fills a critical gap by allowing models to connect past performance with present challenges, fostering a deeper understanding of students' learning behaviors.

3. \textbf{Multi-Agent Collaborative Framework:}  
   To enhance the effectiveness of error analysis and personalized feedback, we propose a novel multi-agent collaborative framework that integrates historical insights with real-time data:
   \begin{itemize}
       \item \textbf{Time Series Agent:} This agent processes students' historical problem-solving data to identify patterns and provide a preliminary classification of errors.
       \item \textbf{MLLM Agent:} Building on the insights of the Time Series Agent, this agent refines error classifications and generates detailed, context-aware suggestions. By combining historical and real-time data, it significantly improves the quality of error analysis and feedback.
   \end{itemize}
   
Together, this work bridges the gap between the capabilities of current AI systems and the expectations of educators. Through \textbf{MathCCS} and our multi-agent collaborative framework, we establish a robust platform for developing intelligent systems that support nuanced reasoning, adaptive learning, and personalized feedback. Our evaluations demonstrate that while current models fall significantly short of human educators, MathCCS provides a critical foundation for advancing research in this domain.

\section{Methodology}

\subsection{Construction of a Multi-modal Error Analysis Benchmark}

\begin{figure}
    \centering
    \includegraphics[width=\linewidth]{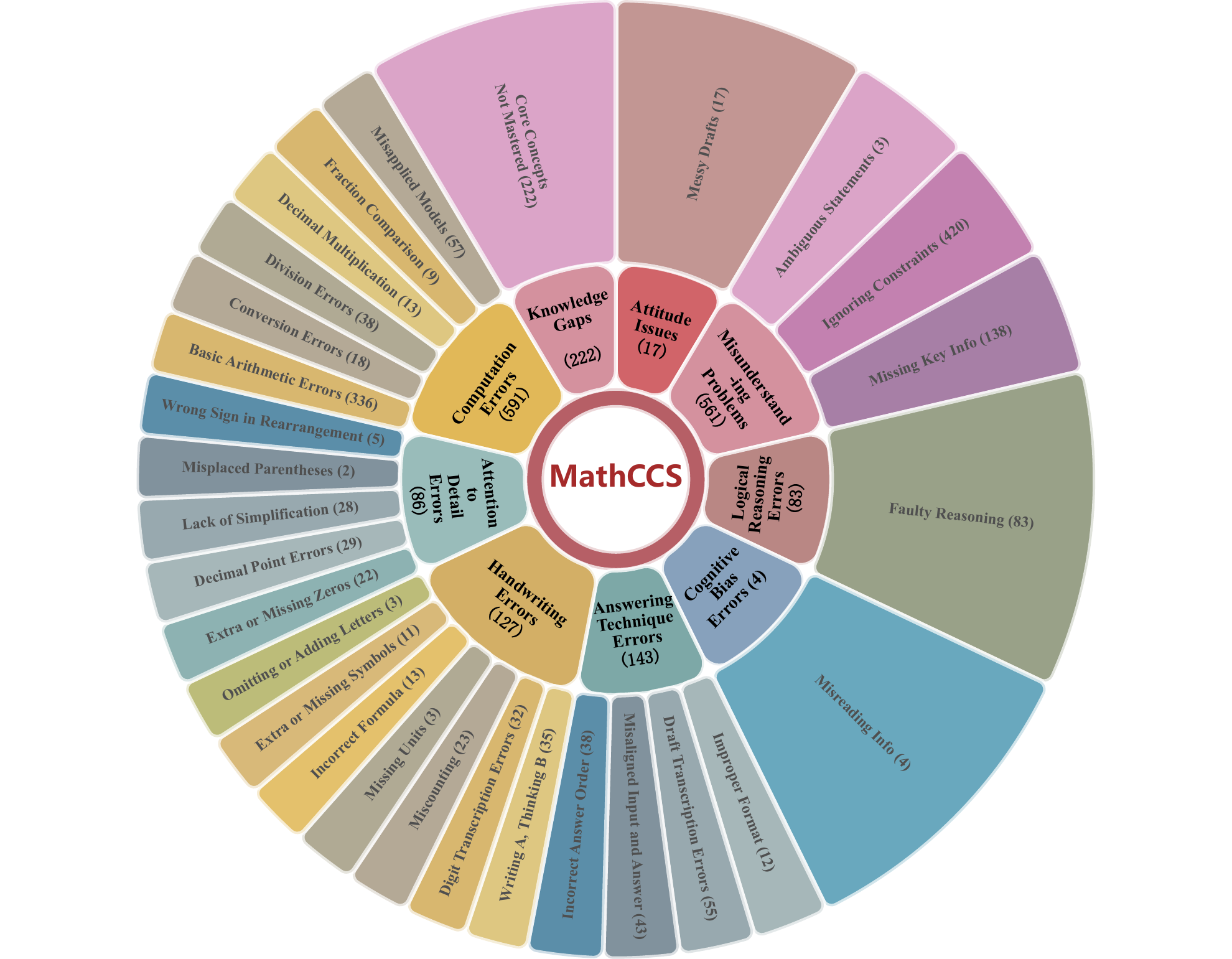}
    \caption{An overview of the MathCCS error categorization framework, showcasing the major error categories and their corresponding subcategories. Detailed explanations for each subcategory are provided in Table~\ref{tab:defi}. The framework is meticulously designed by educational experts, comprising 9 major error categories and 29 subcategories, covering the most prevalent error types observed among elementary-grade students.}
    \label{fig:mathcss}
\end{figure}

To address the limitations of current evaluation systems, \textbf{MathCCS} is introduced as the first multi-modal, multi-type error analysis benchmark tailored for real-world educational challenges. MathCCS provides a comprehensive dataset that integrates realistic problem-solving scenarios, detailed annotations, and personalized improvement suggestions. As shown in Figure~\ref{fig:mathcss}, the benchmark systematically categorizes errors into major categories and subcategories, offering a robust foundation for error analysis and personalized feedback in AI-driven education.

The MathCCS benchmark is built upon three core components to ensure that it captures the complexities of real-world student learning and error patterns:
\begin{enumerate}
    \item \textbf{Real-World Mathematical Problems:} MathCCS incorporates representative mathematical problems drawn directly from real-world educational settings, moving beyond artificial or synthetic benchmarks. Each problem is accompanied by analyses, student responses, and drafts, enabling a deeper understanding of student reasoning and error patterns.

    \item \textbf{Unique Student IDs and Time Stamps:} Each student is assigned a unique identifier, with their responses recorded alongside detailed timestamps. This structure facilitates longitudinal analysis, allowing the construction of user profiles and enabling the evaluation of sequential learning patterns.

    \item \textbf{Categorization of Error Types and Personalized Suggestions:} MathCCS employs a taxonomy of errors defined by educational experts, encompassing four major categories and 37 subcategories. Unlike prior efforts that rely on automated annotations using models like GPT-4, MathCCS annotations are curated by experienced educators to ensure reliability and quality. For each subcategory, detailed suggestions for improvement are included, enabling actionable insights for personalized feedback.
\end{enumerate}

\subsubsection{Data Collection and Filtering}

\textbf{Data Collection:}  
Conventional benchmarks often rely on datasets like GSM8K or other mathematical reasoning datasets, which are limited by synthetic constructs or potential data leakage from model training corpora. MathCCS circumvents these issues by introducing an entirely new dataset built in collaboration with educational experts. The dataset consists of 70,000 problem-solving records collected from elementary-grade students through educational platforms. Each record includes a sequence of problem-solving attempts along with corresponding student drafts, ensuring the dataset's authenticity and applicability to real-world educational contexts.

\textbf{Data Cleaning:}  
To ensure data quality and diversity, a two-step filtering process is applied:
\begin{itemize}
    \item \textbf{Draft Quality Assessment:} GPT-4o evaluates the quality of student drafts based on six key criteria (which is detailed in appendix Table.~\ref{tab:draft_cri}): completeness of steps (30\%), layout and clarity (25\%), correctness of reasoning (20\%), logical consistency (15\%), unit labeling and answer presentation (5\%), and calculation accuracy (5\%). Drafts scoring below 50 are excluded to maintain a high-quality dataset.
    \item \textbf{Diversity Filtering:} To construct a diverse benchmark, 3-gram text-based de-duplication is performed, followed by clustering using the \texttt{all-MiniLM-L6-v2}\footnote{\url{https://huggingface.co/sentence-transformers/all-MiniLM-L6-v2}} sentence embedding model. The embeddings are grouped into 50 clusters using K-means, and 200 samples are randomly selected from each cluster to create a balanced and representative initial dataset.
\end{itemize}

\subsubsection{Data Pre-classification and Reasoning}

\textbf{Predefined Error Types:}  
To establish a robust error taxonomy, a team of experts systematically categorizes common student errors into a hierarchical structure, as shown in Figure~\ref{fig:mathcss}. These predefined categories provide the foundation for annotation and model evaluation, ensuring a comprehensive coverage of error types.

\textbf{Data Pre-alignment:}  
Given the high cost of manual annotation, a pre-classification step is conducted using GPT-4o on 10,000 samples. The model assigns each student's response to one of the predefined error categories and generates preliminary explanations and suggestions. While not perfectly accurate, this pre-classification enables efficient sampling of 100 instances per category, ensuring balanced coverage while reducing the workload for human annotators.

\subsubsection{Annotation and Analysis}

Professional educators annotate 3,700 samples from the pre-aligned dataset, providing:
\begin{itemize}
    \item The specific error category and subcategory.
    \item A detailed explanation of the error's root cause.
    \item Tailored suggestions for improvement, addressing the identified error.
\end{itemize}

The annotation process results in a curated dataset of 1,834 samples. This annotated dataset forms the core of MathCCS, offering a reliable benchmark for evaluating MLLMs in error analysis and personalized feedback generation. MathCCS establishes a critical benchmark for advancing AI-driven education, providing a structured and scalable framework for evaluating error analysis, nuanced reasoning, and personalized feedback in multi-modal educational contexts.

\subsection{Sequential Error Analysis Dataset}
\begin{figure*}[t]
\begin{minipage}[t]{0.51\linewidth}
\centering
 \includegraphics[width=\linewidth]{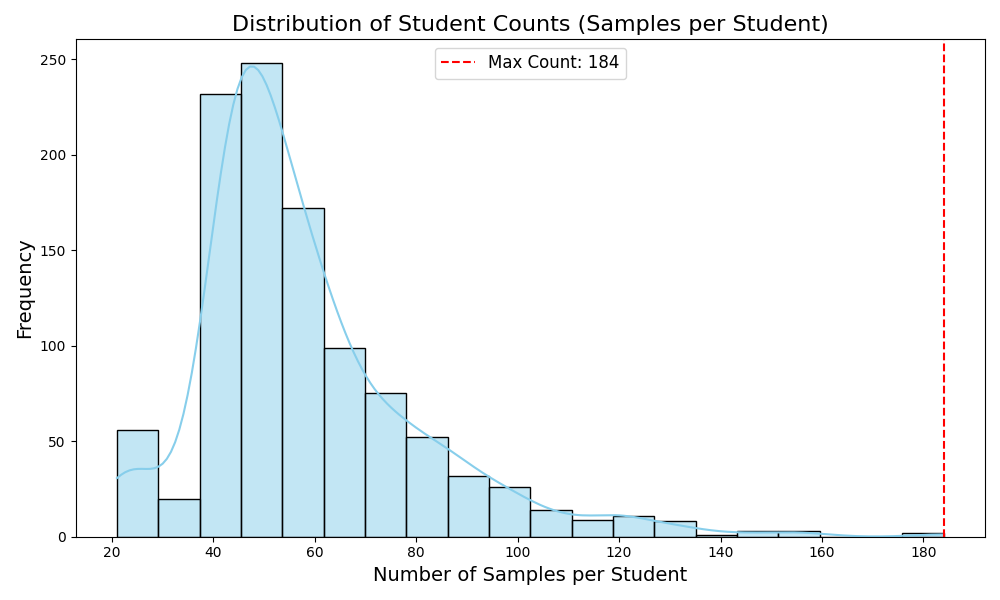}
    \label{fig:phone}
\end{minipage}%
\begin{minipage}[t]{0.45\linewidth}
 \includegraphics[width=\linewidth]{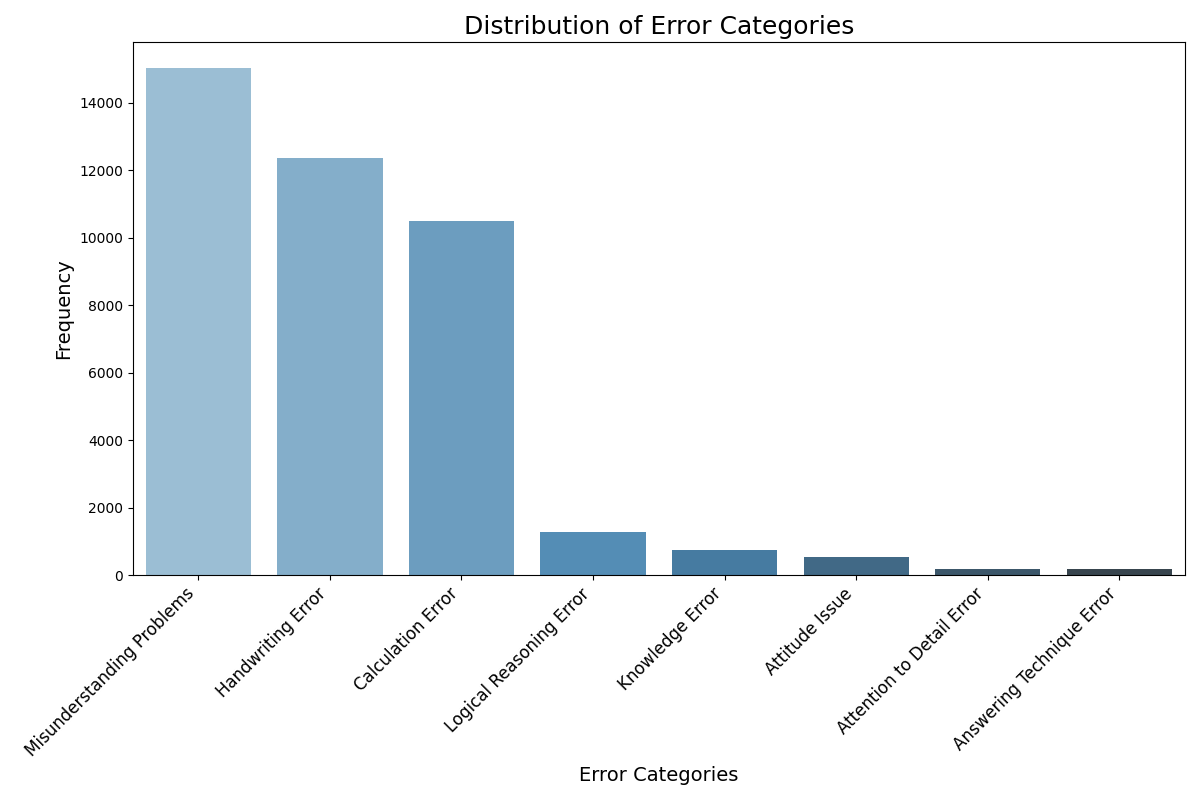}
    \label{fig:adver}
\end{minipage}%
\caption{Left: \textbf{Distribution of student counts in sequential data}, representing the number of data points per student. The minimum value is 21, and the maximum value is 184, showcasing a diverse and realistic dataset of student interaction records in educational scenarios. Right: \textbf{Distribution of error categories identified by GPT-4O in the dataset}. The data exhibits a pronounced long-tail pattern, highlighting the uneven frequency of different error types.}
\label{fig:ocr_cc_perception}
\end{figure*}
The aforementioned benchmarks focus on evaluating, analyzing, and providing recommendations for students' performance on individual tasks. However, a student's learning trajectory evolves over time. For instance, if a student repeatedly makes errors due to lack of attention (e.g., calculation mistakes caused by imprecise draft work), it is reasonable to expect that, following proper guidance and correction, the student should become more sensitive to this type of error in the short term, reducing the likelihood of recurrence. Furthermore, if a student has never encountered problems related to ``cosine functions'', the error rate and reasons for mistakes during the first encounter are likely to differ significantly from those after repeated exposure to similar problems. This illustrates the importance of extracting a student's historical learning profile for analyzing the root causes of errors and providing tailored recommendations. Modeling these dynamics is essential for understanding the temporal evolution of a student's learning behavior and improving adaptive learning systems. However, systematically studying the temporal aspects of student learning presents two major challenges:

\textbf{Data Scarcity} Unlike previous benchmarks, which only require random sampling of students (where approximately 2,000 samples are sufficient to evaluate a model's diagnostic and recommendation capabilities), modeling temporal dependencies necessitates a much larger dataset. First, each student must have an extensive learning record to capture meaningful temporal evolution. This imposes high requirements on individual data collection. Second, it is essential to sample diverse students with long-term records to ensure sufficient variability in the dataset. For instance, a dataset containing only 200 students, each with 50 learning records, would require at least 10,000 samples to support such analysis.

\textbf{Annotation Difficulties} Manually annotating all temporal data, as done in previous benchmarks, is prohibitively costly. This creates a significant bottleneck in terms of annotation resources. To address these challenges, we constructed a dataset comprising 1,063 student samples, each with at least 20 problem-solving records, resulting in a total of 40,772 learning samples. As shown in Figure~\ref{fig:adver} (left), the data originates from real-world problem-solving scenarios involving younger students, making it highly practical and valuable. This dataset fills the current gap in the field of AI-driven education, where temporal data is notably scarce.

For each data point, we employed GPT-4o for annotation. The samples in the training data resemble those in Figure~\ref{fig:data} but do not include error feedback. While our experimental results (detailed in the Results section) validate the differences between GPT-4o annotations and manual annotations, cost constraints necessitated the use of multimodal large language models (MLLMs) as annotators in the current version. In future work, we plan to improve data quality by leveraging more advanced MLLMs or a hybrid approach combining human and MLLM annotations. Despite potential quality issues, it is important to note that GPT-4o does not uniformly categorize errors under the same type, as shown in Figure~\ref{fig:adver} (right). This indicates that, even with lower annotation quality, the dataset retains sufficient diversity. Consequently, this does not impede our research on whether temporal patterns exist within the data. Assuming GPT-4o is a reliable "educator," the dataset can be used to investigate the distributional changes of error types over extended learning periods, enabling the development of more reliable diagnostic tools and recommendations.

\subsection{Multi-Agent Collaborative Framework} 
\begin{figure*}
    \centering
    \includegraphics[width=\linewidth]{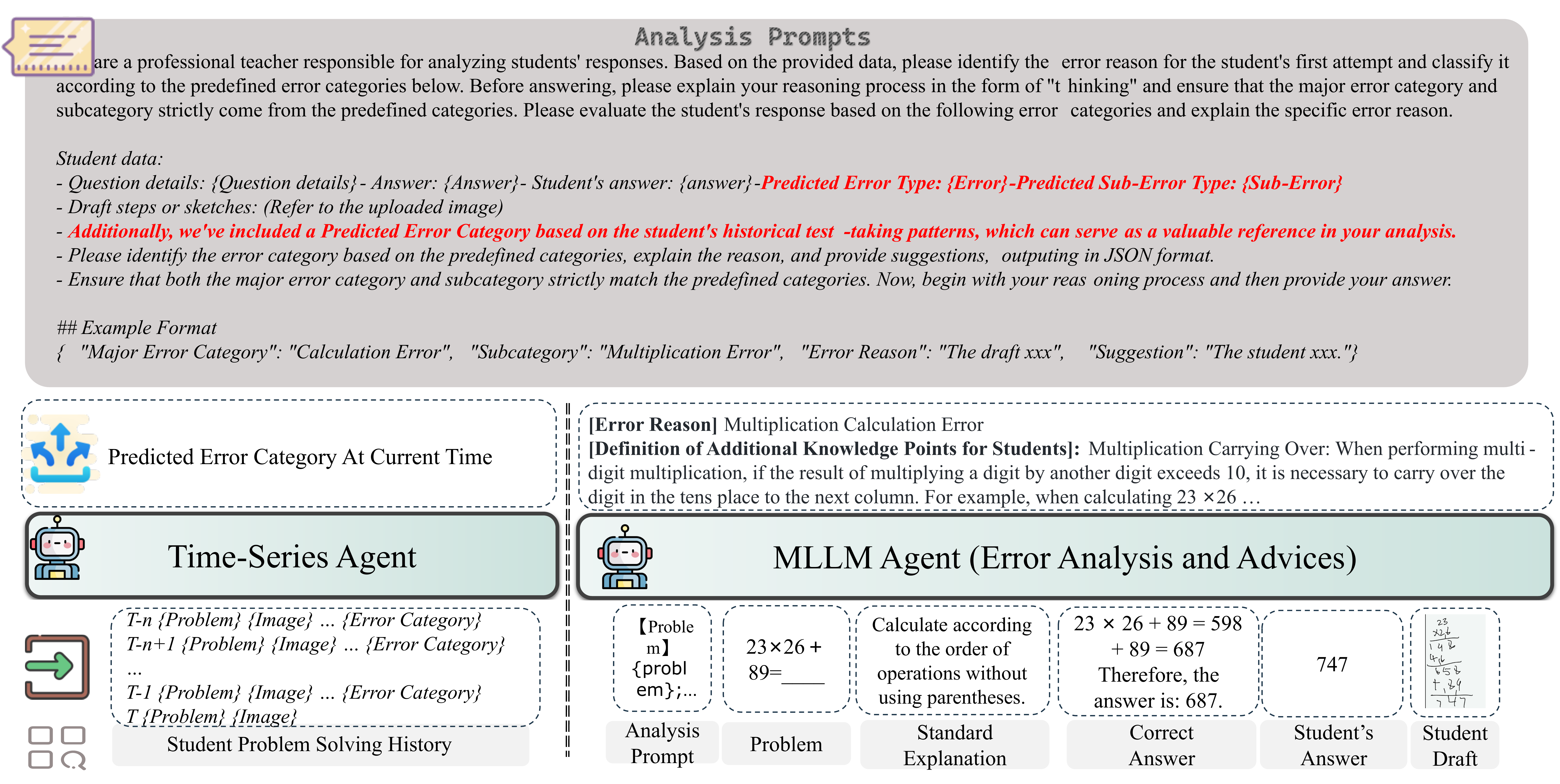}
\caption{The collaborative framework leverages two agents to analyze and understand student problem-solving patterns, along with error diagnosis and recommendations. The Time Series Agent processes historical data on the student's problem-solving behavior to make initial predictions. These preliminary insights are then refined by the MLLM Agent, which employs advanced reasoning capabilities to provide detailed error classifications and context-specific recommendations for improvement. \textbf{The red-highlighted interface represents the output of the time-series model, which is passed to the MLLM agent for downstream error classification and reasoning. If the MLLM's performance is evaluated on individual sample points without leveraging the temporal context, this part of the interface is not required.}}
    \label{fig:model}
\end{figure*}
To enhance the effectiveness of error analysis and personalized feedback, we introduce a multi-agent collaborative framework, as shown in Figure~\ref{fig:model}. This system allocates tasks among multiple intelligent agents, enabling them to work together to analyze student errors from diverse perspectives. By leveraging this multi-agent approach, we can achieve higher accuracy in error diagnosis and provide more nuanced, customized suggestions. Our multi-agent collaborative framework primarily consists of two components:
\subsubsection{Time Series Agent}
\begin{figure}
    \centering
    \includegraphics[width=0.8\linewidth]{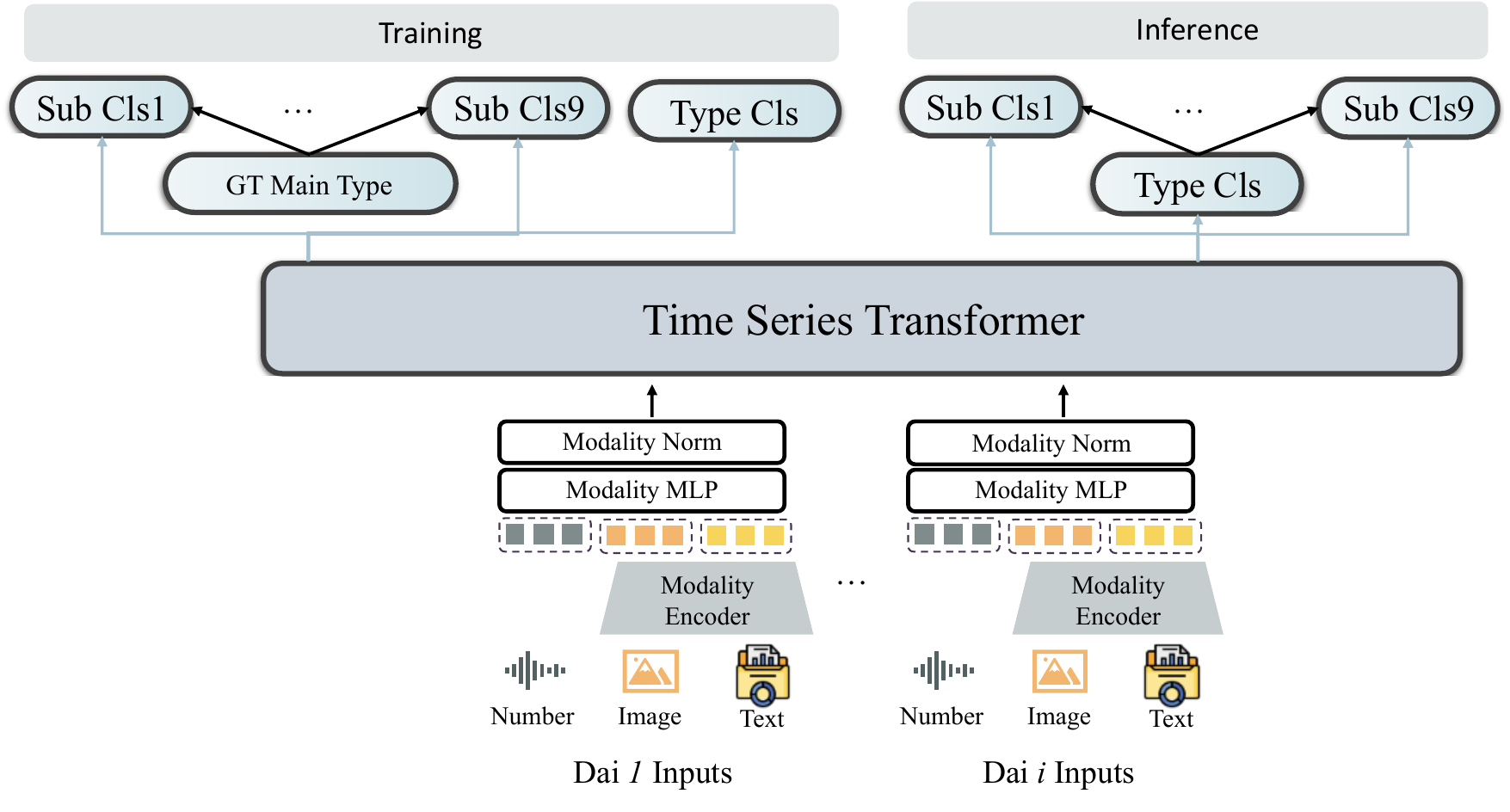}
    \caption{The time-series model architecture consists of modality-specific encoders, an MLP mapping layer, and a pre-normalization layer as the input processing module, which aligns data from different modalities before feeding it into the time-series transformer layer. }
    \label{fig:ts-model}
\end{figure}
The time series agent specializes in processing historical student data. When a student attempts a problem for the $T$th time, the agent takes as input the student's $n$ previous attempts (where $n < T$), including the problem statements, correct answers, the student's responses, draft work, and error types. This historical information, along with the current attempt, is integrated to make an initial prediction. The time series agent synthesizes past data to identify patterns and provide a preliminary assessment.

The design of the time series agent involves the following key requirements:

1. \textbf{Handling Long Context Length and Temporal Features}: The agent must process a large amount of input information over many days, which imposes strict requirements on both context length and temporal order.
2. \textbf{Processing Multi-Modal Inputs}: The input includes not only numerical features like timestamps and student grades but also textual features like problem statements and answers, as well as image features like draft work.

These requirements make it challenging to directly initialize training using existing open-source checkpoints. Current MLLMs often struggle with error classification for a single student sample and are further hindered by the extended context length required to capture meaningful information. Instead, time series models, which typically have fewer temporal parameters and excel at capturing temporal relationships among features~\cite{zhang2024logora,wen2024onenet,nie2022time}, better suit our needs. However, traditional time series models cannot handle multi-modal inputs, and the distribution gap between text, image, and numerical features adds significant complexity. To address this, we propose improvements in both feature input and time series model design (\figurename~\ref{fig:ts-model}):

\paragraph{Feature Input}
There are various types of input for time-series models. Among them, image features include student drafts, text features mainly consist of the question, answer, question explanation, error category, sub error category, and numerical features primarily refer to the timestamp feature. To extract features from both text and images, we utilize CLIP-based models. CLIP, which is trained on large-scale image-text pairs, provides an initial alignment between textual and visual features. However, we observed that time-series models based on CLIP features perform suboptimally, likely due to the relatively low quality of CLIP's training data. To address this, we employ the following approach:

We utilize CLIP-based models to extract features from text and images. CLIP, trained on large-scale image-text pairs, provides preliminary alignment between textual and visual features. However, we found that time series models based on CLIP features perform poorly, likely due to the low quality of CLIP's training data. Thus, we adopted the following approach:
\textbf{For image features}, we use an existing MLLM (default: LLaVA-1.5-7B) to extract features, generating $N_i$ image tokens. To compress the information space further, we process these tokens through the MLLM's language model and use the final token as the input for the time series model:
$F_{\text{image}} = \text{MLLM}_{\text{lang}}(\text{MLLM}_{\text{image}}(I))[-1]$, where $I$ represents the input image, and $F_{\text{image}}$ is the final image token used as the time series model input.

\textbf{For textual features}, a similar process is followed. The text embedding is first generated, passed through the MLLM's language model, and the final token is extracted:
$
F_{\text{text}} = \text{MLLM}_{\text{lang}}(\text{Embed}_{\text{text}}(T))[-1]
$
, where $T$ represents the textual input, and $F_{\text{text}}$ is the condensed textual feature. This approach effectively combines the alignment capabilities of MLLMs with the temporal modeling strength of time series models.

\paragraph{Time Series Model Design}
\textbf{Pre-Norm and alignment module} The main framework is similar to PatchTST~\cite{nie2022time}, a widely validated method in time series research. PatchTST assumes independence among different channels of time series data, allowing for separate predictions. As shown in \figurename~\ref{fig:ts-model}, we enhance this method by adding modality-specific normalization and linear layers to align features across modalities before input.
The overall input representation for a time series sample at time $t$ can be expressed as:
$
X_t = \text{Align}(\text{Norm}_{\text{modality}}(F^{\text{img}}, F^{\text{txt}}, F^{\text{num}})),
$
where $\text{Norm}_{\text{modality}}$ and $\text{Align}$ denote modality-specific normalization and alignment layers, respectively. For classification tasks, the time series model predicts both error types ($y^{\text{type}}$) and subtypes ($y^{\text{subtype}}$). These tasks are interdependent. To model this dependency effectively, we use a teacher-forcing approach during training. The error type classifier predicts $y^{\text{type}}$, and its ground truth is used to guide the subtype classification. The loss function is given as:
\begin{equation}
\mathcal{L} = \mathcal{L}_{\text{type}}(y^{\text{type}}, \hat{y}^{\text{type}}) + \mathcal{L}_{\text{subtype}}(y^{\text{subtype}}, \hat{y}^{\text{subtype}} | y^{\text{type}}),
\end{equation}
where $\mathcal{L}_{\text{type}}$ and $\mathcal{L}_{\text{subtype}}$ are the respective loss functions for type and subtype classification.

During inference, the predicted error type $\hat{y}^{\text{type}}$ is directly used to activate the subtype classifier:
\begin{equation}
\hat{y}^{\text{subtype}} = \text{SubtypeClassifier}(\hat{y}^{\text{type}}, X_t).
\end{equation}

This design mitigates error propagation during training and improves the robustness of subtype classification during inference.

\subsubsection{MLLM Agent}

While the predictions from the time series agent serve as an initial assessment, they may be limited due to fewer training parameters and the inability to generate detailed, coherent textual suggestions. To address this, we developed a second component—the MLLM (Multi-Modal Large Language Model) agent. This agent refines the initial predictions from the time series agent by incorporating insights from the student's current response, offering more precise error classification and generating detailed, context-specific recommendations.

\section{Experiments}

\subsection{Performance Analysis of MLLMs on MathCCS}

\subsubsection{Evaluation Metric}  
First, we define all error categories and subcategories for MLLMs and use the following prompt to request results. The command “Please explain your reasoning process in the form of 'thinking'” ensures that the model thinks in a CoT (Chain of Thought) format. We found that this step makes the results more accurate (the detailed prompt is similar to Fig.~\ref{fig:model} but excludes the contents of the Predicted Error Type).  
\textbf{Evaluation Standards:}  
\begin{enumerate}
    \item \textbf{Accuracy of Error Classification:}  
    For the major and subcategories of errors, we directly compare the model's output against the predefined categories.  
    The classification task is evaluated using accuracy:  
    $
    \text{Accuracy} = \frac{\text{Number of Correct Classifications}}{\text{Total Classifications}} \times 100\%
    $

    \item \textbf{Relevance of Error Reasons:}  
    For error reasons and suggestions, we re-query GPT-4o to compare the model's outputs with those provided by human experts. GPT-4o scores the outputs on a scale of 0 to 10, where:  
        a) \textbf{0-2:} Completely mismatched; the reason or suggestion deviates entirely from the context or fails to explain the error.
        b) \textbf{3-5:} Partially correct; the output captures some aspects of the reasoning but lacks key details or precision.
        c) \textbf{6-8:} Nearly correct; the output aligns closely with the expert's reasoning but may miss minor details or nuances.
        d) \textbf{9-10:} Fully correct; the output matches the human expert's reasoning perfectly in both content and context.

    \item \textbf{Actionability of Suggestions:}  
    Compare the model's suggestions with those provided by experts, focusing on their practicality and specificity. GPT-4o scores the actionability of suggestions on the same 0-10 scale.  
\end{enumerate}

\subsubsection{Experimental Results}  
Our experiments reveal significant limitations in the current generation of MLLMs regarding their ability to classify student errors and provide actionable suggestions. Below, we summarize the primary observations and challenges:  

\begin{enumerate}
    \item \textbf{Limited Error Classification Capability:}  
    Certain models, such as LLaVA-OV-7B and LLaMA 3.2-8B, heavily rely on oversimplified categorization strategies, often misclassifying all types of errors as ``Computation Errors". This behavior indicates a lack of nuanced understanding of error types, such as ``Misunderstanding" or ``Attention to Detail Errors".  

    \item \textbf{Low Average Classification Accuracy:}  
    Even the most advanced models, including Claude3.5-Sonnet and GPT-4o, achieve an average classification accuracy below 30\%. This is substantially lower than the performance of human experts, who can distinguish between subtle differences in error categories with a much higher degree of accuracy.  

    \item \textbf{Significant Gap Between MLLMs and Human Experts:}  
    The current models exhibit a marked inability to provide reliable diagnostic insights across multiple error types. Categories such as ``Knowledge Gaps" and "Logical Reasoning" are particularly challenging for these models. This demonstrates that existing MLLMs lack the contextual understanding and reasoning capabilities required for robust error classification.  

    \item \textbf{Evaluation Metrics Summary:}  
    \begin{itemize}
        \item \textbf{Classification Accuracy:} Models perform well below human levels, with an average accuracy of 10-30\% on complex error categorization tasks.  
        \item \textbf{Relevance Scores (0-10):} Model-generated error reasons score between 1-4, showing moderate alignment with human outputs but lacking depth in complex cases.  
        \item \textbf{Actionability Scores (0-10):} Suggestions generated by models score between 1-3, indicating significant limitations in practical and actionable advice compared to humans.  
    \end{itemize}
\end{enumerate}

The results highlight the current limitations of MLLMs in educational contexts, especially in tasks requiring detailed error analysis and actionable suggestions. While these models may assist in preliminary assessments or repetitive tasks, they are far from replacing the expertise of human teachers. Professional educators remain indispensable for providing context-aware feedback and personalized instruction.

\begin{table}[H]
\centering
\caption{Comparison of seven models across three evaluation metrics: \textbf{Accuracy}, \textbf{Average Reasoning Score}, and \textbf{Average Suggestion Score}. Each subtable presents a separate metric for all models, arranged horizontally for comparison.}
\label{tab:model_performance}
\begin{adjustbox}{width=0.95\textwidth}
\begin{tabular}{lccccccc}
    \toprule
    \textbf{Category} & \textbf{Claude3.5-Sonnet} & \textbf{GPT-4o-Mini} & \textbf{GPT-4o} & \textbf{Qwen2-VL-7B} & \textbf{InternVL 2.5-8B}& \textbf{LLaVA-OV-7B} &  \textbf{LLaMA 3.2-8B}   \\
    \midrule
    \textbf{Accuracy} \\
    \midrule
Computation Errors & 0.6373 & 0.8003 & 0.0180 & 0.9222 & 0.0856 & 0.9983 & 0.8985 \\
Misunderstanding & 0.1889 & 0.1800 & 0.4602 & 0.0089 & 0.8596 & 0.0000 & 0.0000 \\
Attention to Detail Error & 0.2471 & 0.0000 & 0.1337 & 0.0000 & 0.0236 & 0.0000 & 0.0000 \\
Handwriting & 0.1181 & 0.0157 & 0.0000 & 0.0000 & 0.0000 & 0.0000 & 0.0000 \\
Knowledge Gaps & 0.1312 & 0.0270 & 0.0964 & 0.0000 & 0.0090 & 0.0000 & 0.0000 \\
Cognitive Bias & 0.0000 & 0.0000 & 0.0866 & 0.0000 & 0.0000 & 0.0000 & 0.0000 \\
Logical Reasoning & 0.0964 & 0.1446 & 0.3217 & 0.1566 & 0.0241 & 0.0000 & 0.0000 \\
Answering Technique & 0.5315 & 0.1329 & 0.1163 & 0.0000 & 0.0000 & 0.0000 & 0.0000 \\
Attitude & 0.1176 & 0.0588 & 0.1176 & 0.0000 & 0.0588 & 0.0000 & 0.0000 \\
Avg & \textbf{0.2298} & 0.1510 & 0.1501 & 0.1209 & 0.1179 & 0.1109 & 0.0998 \\
    \midrule
    \textbf{Average Reasoning Score} \\
    \midrule
Computation Errors & 2.7102 & 2.5262 & 4.1118 & 1.6684 & 2.1105 & 0.4247 & 0.8646 \\
Misunderstanding & 4.4385 & 2.9607 & 3.0752 & 1.3529 & 2.0423 & 0.6631 & 0.3119 \\
Attention to Detail Error & 3.0118 & 1.9070 & 4.0301 & 0.9651 & 1.3780 & 0.0814 & 1.5814 \\
Handwriting & 3.4646 & 1.9528 & 4.0000 & 1.2283 & 1.5349 & 0.2441 & 0.4646 \\
Knowledge Gaps & 3.8688 & 3.7658 & 3.5614 & 2.0225 & 2.8694 & 0.8604 & 0.4459 \\
Cognitive Bias & 1.5000 & 1.5000 & 3.6623 & 1.0000 & 1.2500 & 0.5000 & 0.2500 \\
Logical Reasoning & 3.7349 & 3.3373 & 5.0917 & 1.6265 & 2.3133 & 0.5542 & 0.2651 \\
Answering Technique & 5.2797 & 2.6504 & 3.8750 & 0.7762 & 1.0629 & 0.0699 & 0.4266 \\
Attitude & 0.1765 & 0.1765 & 0.5000 & 0.2353 & 0.0588 & 0.2353 & 0.1765 \\
Avg & 3.1317 & 2.3085 & \textbf{3.5453} & 1.2084 & 1.6244 & 0.4037 & 0.5318 \\
    \midrule
    \textbf{Average Suggestion Score} \\
    \midrule
Computation Errors & 1.9881 & 1.7885 & 2.6579 & 1.5601 & 1.8877 & 0.6430 & 0.6920 \\
Misunderstanding & 3.2567 & 2.3161 & 2.1671 & 1.6453 & 1.6954 & 1.4421 & 0.6221 \\
Attention to Detail Error & 2.3412 & 1.6395 & 3.0115 & 1.3023 & 1.6063 & 0.3023 & 1.5349 \\
Handwriting & 2.1654 & 1.6772 & 1.5000 & 1.4173 & 1.8605 & 0.5748 & 0.8032 \\
Knowledge Gaps & 2.4027 & 2.4189 & 2.0351 & 1.2117 & 1.7838 & 0.8423 & 0.2658 \\
Cognitive Bias & 1.2500 & 0.0000 & 2.9351 & 0.5000 & 0.5000 & 0.7500 & 0.7500 \\
Logical Reasoning & 2.4699 & 1.9518 & 3.8440 & 1.3494 & 1.4940 & 0.6145 & 0.1325 \\
Answering Technique & 3.5035 & 1.7692 & 3.4219 & 1.2098 & 1.2937 & 0.3986 & 0.8252 \\
Attitude & 0.4706 & 0.4706 & 0.2500 & 0.2941 & 0.2941 & 0.2353 & 0.2941 \\
Avg & 2.2053 & 1.5591 & \textbf{2.4247} & 1.1656 & 1.3795 & 0.6448 & 0.6578\\
    \bottomrule
\end{tabular}
\end{adjustbox}
\end{table}

\subsection{Sequential Error Analysis}

\subsubsection{Experiment Setup}
For the time series transformer, we adopt the PatchTST architecture~\cite{nie2022time} but remove its classification head and adjust its model size. The default architecture uses 6 layers with a hidden dimension and MLP mapping size of 256, and an attention head size of 8, resulting in fewer than 20 million parameters. The learning rate, determined via grid search, is set to $5\times10^{-4}$, with minimal impact on model optimization. The time series data described earlier is split into training, validation, and test sets in an 8:1:1 ratio. The maximum training epochs are set to 40, with an early stopping patience of 3, meaning training halts if the validation loss does not improve for three consecutive epochs. The checkpoint with the lowest validation loss is retained as the final model.

To construct student time-series data, we utilize historical records consisting of up to 20 problem-solving attempts per student. These 20 historical records, along with their associated error classifications, are used as input features to predict the error classification of the student’s 21st attempt. To ensure the reliability and consistency of the student profiles, only records where the time interval between two consecutive attempts is less than one month are included. If the time gap exceeds one month, the student profile is deemed unreliable, and such data is excluded from both the training and evaluation datasets. This filtering approach ensures the temporal consistency of the input data, thereby improving the quality and robustness of the model's predictions.

\subsubsection{Designing the Time Series Model}

Table~\ref{tab:methods_accuracy} evaluates various design choices:

- \textbf{Baseline}: Measures performance without considering sequential information.

- \textbf{Modality Input}: Explores strategies for processing multimodal data.

- \textbf{Pre-Norm}: Examines the effect of modality-specific normalization before feature processing.

- \textbf{Type Weight}: Investigates the use of class-specific weights in the cross-entropy loss to address the long-tail problem.

- \textbf{Classifier}: Compares different classifier designs, including:
  - \textbf{F-Conditioned}: Incorporates user features into subtype classification logits.
  - \textbf{I-Conditioned}: Activates distinct subtype classifiers based on error type predictions.
  - \textbf{Teacher-force}: Implements the teacher-forcing method introduced in the main text.
  
- \textbf{Token for Classifier}: Analyzes different token choices for classifier input.

\paragraph{Key Findings}

1. \textbf{Multimodal Feature Processing}: Using raw CLIP features produces suboptimal results. Stronger MLLMs provide better image and text features. However, directly using all MLLM tokens introduces excessive redundancy (nearly 1,000 tokens per sample), which small-scale time series models cannot handle efficiently. Aggregating features with $\text{MLLM}_{\text{lang}}$ and selecting only the final token achieves the best performance.

2. \textbf{Pre-Normalization}: Adding pre-normalization is essential for effectively integrating multimodal features.

3. \textbf{Addressing Long-Tail Distributions}: Given the evident long-tail distribution in the data, weighting cross-entropy loss by class sample counts slightly improves performance for minority classes. However, this approach significantly degrades the performance of the majority classes.

4. \textbf{Classifier Design}:
   - Independent error type and subtype classifiers perform poorly, especially for subtypes.
   - Conditioning subtype classification logits on error type predictions (F-Conditioned) improves subtype performance but introduces error accumulation, degrading error type results.
   - The I-Conditioned method decouples the gradients of error type and subtype classifiers while activating subtype classifiers based on error type predictions. Combined with teacher-forcing training, this approach yields superior results.

5. \textbf{Token Selection for Classification}: Using the mean of token embeddings performs well for error type classification but lacks fine-grained subtype classification capability. Adding a [CLS] token at the input sequence's end and concatenating it with the averaged features as classifier input achieves optimal results.

\begin{table}[ht]
\centering
\caption{\textbf{Comparison of sequential model designs and their impact on accuracy} (Acc1 for error type classification, Acc2 for subtype classification). }
\label{tab:methods_accuracy}
\renewcommand{\arraystretch}{1.2} 
\begin{adjustbox}{width=0.85\textwidth}
\begin{tabular}{@{}llcc|llcc@{}}
\toprule
\multicolumn{4}{c|}{\textbf{Method}} & \multicolumn{4}{c}{\textbf{Method}} \\ 
\cmidrule(r){1-4} \cmidrule(l){5-8}
\textbf{Main Type} & \textbf{Sub Type} & \textbf{Acc1} & \textbf{Acc2} & \textbf{Main Type} & \textbf{Sub Type} & \textbf{Acc1} & \textbf{Acc2} \\ 
\midrule
Baseline         & wo. sequential data & 0.20 & 0.13 &        \multirow{4}{*}{Classifier}   & Individual       & 0.28 & 0.09 \\
\cmidrule{1-4}
\multirow{3}{*}{Modality Input} 
                 & CLIP Encoder        & 0.23 & 0.21 & & F-Conditioned              & 0.27 & 0.23 \\
                 & LLaVA-1.5 Encoder   & 0.29 & 0.31 &                       & I-Conditioned   & 0.33 & 0.33 \\
                 & LLM Pooling         & 0.34 & 0.43 &                       & Teacher-force   & 0.34 & 0.43 \\
\midrule
\multirow{2}{*}{Pre Norm} 
                 & wo. Norm            & 0.24 & 0.25 & \multirow{4}{*}{Token for Classifier} 
                                                                                       & Last Token     & 0.31 & 0.33 \\
                 & w. Norm             & 0.34 & 0.43 &                       & Mean            & 0.36 & 0.20 \\
\cmidrule{1-4}
\multirow{2}{*}{Type Weight} 
                 & wo. type weight     & 0.34 & 0.43 &                       & [CLS] Token     & 0.31 & 0.35 \\
                 & w. type weight      & 0.30 & 0.28 &                       & Mean + [CLS]    & 0.34 & 0.43 \\
\bottomrule
\end{tabular}
\end{adjustbox}
\end{table}
\subsubsection{Integrating Time-Series Agents with MLLM Agents}

\begin{table}[ht]
\centering
\captionsetup{font=small, labelfont=bf}
\caption{Performance Comparison of MLLM Agents}\label{tab:multi_agent}
\begin{adjustbox}{width=0.85\textwidth}
\begin{tabular}{cccccc}
\toprule
\textbf{MLLM+Time Agent} & \textbf{Model} & \textbf{Acc-Type} & \textbf{Acc-Subtype}& \textbf{Reasoning Score} & \textbf{Suggestion Score} \\
\midrule
\multirow{2}{*}{w/o. time agent} & InternVL2.5-8B & 0.1686 & 0.0636 & 1.9113 & 1.6704 \\
 & Qwen2-VL-8B & 0.1835 & 0.0753 & 2.3741 & 1.9873 \\
\midrule
\multirow{2}{*}{w. time agent}  & InternVL2.5-8B & 0.2497 & 0.2112 & 2.3842 & 2.1858 \\
 & Qwen2-VL-8B & 0.2632 & 0.1619 & 2.9622 & 2.7641 \\
\bottomrule
\end{tabular}
\end{adjustbox}
\end{table}
From the results presented in the previous section, it is clear that task-specific time-series agents exhibit strong performance in error classification. While the benchmark data used here is not identical to MathCSS, it is noteworthy that, in MathCSS—where the accuracy of the most advanced MLLMs has yet to surpass 30\%—time-series agents with relatively modest sizes (ranging from tens to hundreds of millions of parameters) can achieve comparable performance after targeted training. However, despite their classification capabilities, time-series agents lack the reasoning and generative functionalities inherent to MLLMs, rendering them unsuitable for independently producing detailed error analyses or actionable recommendations based on their predictions.

In this section, we explore the integration of the best-trained time-series agents from the previous section with two of the most advanced MLLM agents, InternVL 2.5-8B and Qwen2VL-8B, to assess whether this hybrid approach can more effectively build comprehensive student profiles and deliver precise, actionable recommendations using historical information. It is important to note that the evaluation data used in this section is not derived from MathCSS but from a test set of sequential data annotated by GPT-4o.

\begin{figure}
    \centering
    \includegraphics[width=\linewidth]{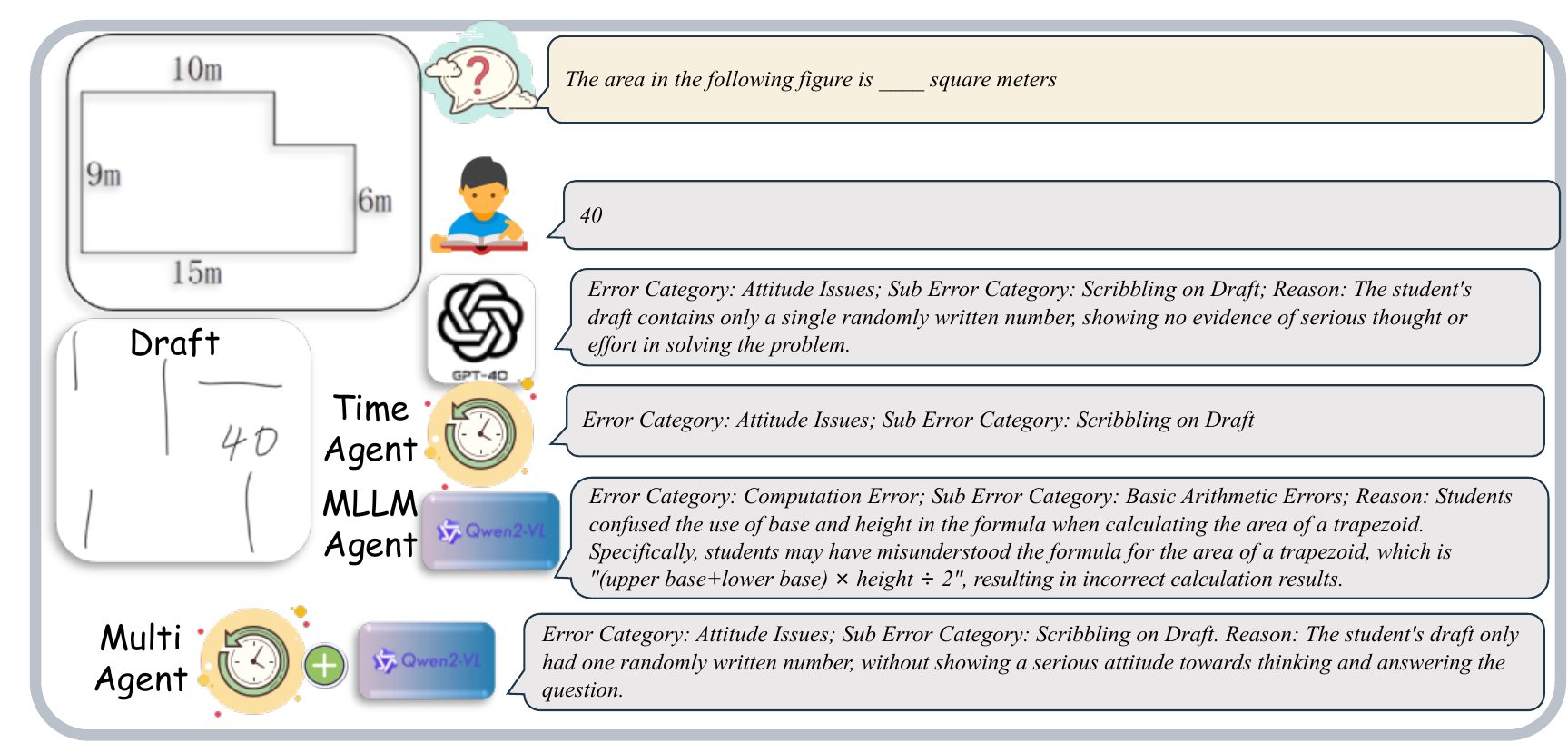}
    \caption{\textbf{Comparison between multi-agent and MLLM agent systems.} The MLLM agent struggles with accurately identifying error types, leading to downstream issues in causal analysis and recommendations. In contrast, the sequential model is task-specific, trained on dedicated datasets, and excels at classification. However, it lacks open-ended generation capabilities, making it unsuitable for interactive student engagement. By combining the strengths of both, the model achieves strong performance in both classification and personalized recommendation.}
\label{fig:multi_agent}
\end{figure}

The results in Table~\ref{tab:multi_agent} indicate that current MLLMs still perform suboptimally in professional error analysis tasks, consistent with their limited performance on MathCSS. However, incorporating time-series agents into the workflow significantly enhances error classification accuracy. Figure~\ref{fig:multi_agent} further illustrates several examples where the inclusion of time-series classification led to more accurate final predictions. Notably, the models do not always directly adopt the predictions of the time-series agent; instead, they selectively reference and adapt the final outputs. This selective integration not only improves prediction success rates but also enables the generation of more coherent and insightful error analyses and recommendations.

Interestingly, there were instances where the time-series agent correctly identified the error type, but the MLLM failed to adjust its output accordingly. We hypothesize that such discrepancies may stem from overconfidence in the MLLM’s predictions in these cases. Looking forward, we propose two potential strategies to address this issue: (1) fine-tuning the MLLM to better integrate and interpret the outputs of the time-series agent, or (2) leveraging the time-series agent as a highly reliable signal, potentially treating its predictions as ground truth for classification tasks as its performance continues to improve.

\section{Conclusion and Future Directions}
Personalized education requires a deep understanding of each student's learning process, including the reasons behind their mistakes and the best way to correct them. Our multi-modal error analysis benchmark and sequential learning dataset provide a foundation for developing models that can offer such insights. By highlighting the limitations of existing systems and proposing a multi-agent collaborative framework, we demonstrate a pathway toward more effective, personalized education that can adapt to the needs of each individual learner.

Our work lays the groundwork for the development of more advanced, personalized educational tools. Future research will focus on enhancing the capabilities of the multi-agent framework, optimizing it for more complex educational tasks. Additionally, we plan to improve the integration of multi-modal data, enabling models to better understand and analyze diverse problem-solving scenarios. By refining these components, we aim to set a new standard for personalized education, where each student's learning journey is acknowledged and supported.



---------------------------------------------------
%

\bigskip
\bibliography{main}

\begin{thebibliography}{10}

\bibitem{awadalla2023openflamingo}
Anas Awadalla, Irena Gao, Josh Gardner, Jack Hessel, Yusuf Hanafy, Wanrong Zhu, Kalyani Marathe, Yonatan Bitton, Samir Gadre, Shiori Sagawa, et~al.
\newblock Openflamingo: An open-source framework for training large autoregressive vision-language models.
\newblock {\em arXiv preprint arXiv:2308.01390}, 2023.

\bibitem{bai2023qwen}
Jinze Bai, Shuai Bai, Shusheng Yang, Shijie Wang, Sinan Tan, Peng Wang, Junyang Lin, Chang Zhou, and Jingren Zhou.
\newblock Qwen-vl: A frontier large vision-language model with versatile abilities.
\newblock {\em arXiv preprint arXiv:2308.12966}, 2023.

\bibitem{bitton2023visit}
Yonatan Bitton, Hritik Bansal, Jack Hessel, Rulin Shao, Wanrong Zhu, Anas Awadalla, Josh Gardner, Rohan Taori, and Ludwig Schimdt.
\newblock Visit-bench: A benchmark for vision-language instruction following inspired by real-world use.
\newblock {\em arXiv preprint arXiv:2308.06595}, 2023.

\bibitem{brown2020language}
Tom Brown, Benjamin Mann, Nick Ryder, Melanie Subbiah, Jared~D Kaplan, Prafulla Dhariwal, Arvind Neelakantan, Pranav Shyam, Girish Sastry, Amanda Askell, et~al.
\newblock Language models are few-shot learners.
\newblock {\em NeurIPS}, 2020.

\bibitem{chiang2023vicuna}
Wei-Lin Chiang, Zhuohan Li, Zi~Lin, Ying Sheng, Zhanghao Wu, Hao Zhang, Lianmin Zheng, Siyuan Zhuang, Yonghao Zhuang, Joseph~E Gonzalez, et~al.
\newblock Vicuna: An open-source chatbot impressing gpt-4 with 90\%* chatgpt quality.
\newblock {\em https://vicuna.lmsys.org}, 2023.

\bibitem{dai2024instructblip}
Wenliang Dai, Junnan Li, Dongxu Li, Anthony Meng~Huat Tiong, Junqi Zhao, Weisheng Wang, Boyang Li, Pascale~N Fung, and Steven Hoi.
\newblock Instructblip: Towards general-purpose vision-language models with instruction tuning.
\newblock {\em NeurIPS}, 2024.

\bibitem{fu2023mme}
Chaoyou Fu, Peixian Chen, Yunhang Shen, Yulei Qin, Mengdan Zhang, Xu~Lin, Jinrui Yang, Xiawu Zheng, Ke~Li, Xing Sun, et~al.
\newblock Mme: A comprehensive evaluation benchmark for multimodal large language models.
\newblock {\em arXiv preprint arXiv:2306.13394}, 2023.

\bibitem{fu2024vita}
Chaoyou Fu, Haojia Lin, Zuwei Long, Yunhang Shen, Meng Zhao, Yifan Zhang, Shaoqi Dong, Xiong Wang, Di~Yin, Long Ma, et~al.
\newblock Vita: Towards open-source interactive omni multimodal llm.
\newblock {\em arXiv preprint arXiv:2408.05211}, 2024.

\bibitem{fu2023challenger}
Chaoyou Fu, Renrui Zhang, Haojia Lin, Zihan Wang, Timin Gao, Yongdong Luo, Yubo Huang, Zhengye Zhang, Longtian Qiu, Gaoxiang Ye, et~al.
\newblock A challenger to gpt-4v? early explorations of gemini in visual expertise.
\newblock {\em arXiv preprint arXiv:2312.12436}, 2023.

\bibitem{fu2024mme}
Chaoyou Fu, Yi-Fan Zhang, Shukang Yin, Bo~Li, Xinyu Fang, Sirui Zhao, Haodong Duan, Xing Sun, Ziwei Liu, Liang Wang, et~al.
\newblock Mme-survey: A comprehensive survey on evaluation of multimodal llms.
\newblock {\em arXiv preprint arXiv:2411.15296}, 2024.

\bibitem{fu2024blink}
Xingyu Fu, Yushi Hu, Bangzheng Li, Yu~Feng, Haoyu Wang, Xudong Lin, Dan Roth, Noah~A Smith, Wei-Chiu Ma, and Ranjay Krishna.
\newblock Blink: Multimodal large language models can see but not perceive.
\newblock {\em arXiv preprint arXiv:2404.12390}, 2024.

\bibitem{gannot2023data}
Sharon Gannot, Zheng-Hua Tan, Martin Haardt, Nancy~F Chen, Hoi-To Wai, Ivan Tashev, Walter Kellermann, and Justin Dauwels.
\newblock Data science education: The signal processing perspective [sp education].
\newblock {\em IEEE Signal Processing Magazine}, 40(7):89--93, 2023.

\bibitem{han2023coremm}
Xiaotian Han, Quanzeng You, Yongfei Liu, Wentao Chen, Huangjie Zheng, Khalil Mrini, Xudong Lin, Yiqi Wang, Bohan Zhai, Jianbo Yuan, Heng Wang, and Hongxia Yang.
\newblock Infimm-eval: Complex open-ended reasoning evaluation for multi-modal large language models, 2023.

\bibitem{li2023blip}
Junnan Li, Dongxu Li, Silvio Savarese, and Steven Hoi.
\newblock Blip-2: Bootstrapping language-image pre-training with frozen image encoders and large language models.
\newblock {\em arXiv preprint arXiv:2301.12597}, 2023.

\bibitem{liu2023visual}
Haotian Liu, Chunyuan Li, Qingyang Wu, and Yong~Jae Lee.
\newblock Visual instruction tuning.
\newblock {\em arXiv preprint arXiv:2304.08485}, 2023.

\bibitem{lu2024mathvista}
Pan Lu, Hritik Bansal, Tony Xia, Jiacheng Liu, Chunyuan Li, Hannaneh Hajishirzi, Hao Cheng, Kai-Wei Chang, Michel Galley, and Jianfeng Gao.
\newblock Mathvista: Evaluating mathematical reasoning of foundation models in visual contexts.
\newblock In {\em ICLR}, 2024.

\bibitem{meier2015predicting}
Yannick Meier, Jie Xu, Onur Atan, and Mihaela Van~der Schaar.
\newblock Predicting grades.
\newblock {\em IEEE Transactions on Signal Processing}, 64(4):959--972, 2015.

\bibitem{nie2022time}
Yuqi Nie, Nam~H Nguyen, Phanwadee Sinthong, and Jayant Kalagnanam.
\newblock A time series is worth 64 words: Long-term forecasting with transformers.
\newblock {\em The Eleventh International Conference on Learning Representations (ICLR'23)}, 2023.

\bibitem{gpt4}
OpenAI.
\newblock Gpt-4 technical report.
\newblock 2023.

\bibitem{prochazka2021integrating}
Ales Prochazka, Oldrich Vysata, and Vladimir Marik.
\newblock Integrating the role of computational intelligence and digital signal processing in education: Emerging technologies and mathematical tools.
\newblock {\em IEEE Signal Processing Magazine}, 38(3):154--162, 2021.

\bibitem{roy2018mapping}
Subhro Roy and Dan Roth.
\newblock Mapping to declarative knowledge for word problem solving.
\newblock {\em Transactions of the Association for Computational Linguistics}, 6:159--172, 2018.

\bibitem{singh2019towards}
Amanpreet Singh, Vivek Natarajan, Meet Shah, Yu~Jiang, Xinlei Chen, Dhruv Batra, Devi Parikh, and Marcus Rohrbach.
\newblock Towards vqa models that can read.
\newblock In {\em CVPR}, 2019.

\bibitem{touvron2023llama}
Hugo Touvron, Thibaut Lavril, Gautier Izacard, Xavier Martinet, Marie-Anne Lachaux, Timoth{\'e}e Lacroix, Baptiste Rozi{\`e}re, Naman Goyal, Eric Hambro, Faisal Azhar, et~al.
\newblock Llama: Open and efficient foundation language models.
\newblock {\em arXiv preprint arXiv:2302.13971}, 2023.

\bibitem{touvron2023llama2}
Hugo Touvron, Louis Martin, Kevin Stone, Peter Albert, Amjad Almahairi, Yasmine Babaei, Nikolay Bashlykov, Soumya Batra, Prajjwal Bhargava, Shruti Bhosale, et~al.
\newblock Llama 2: Open foundation and fine-tuned chat models.
\newblock {\em arXiv preprint arXiv:2307.09288}, 2023.

\bibitem{wang2024large}
Shen Wang, Tianlong Xu, Hang Li, Chaoli Zhang, Joleen Liang, Jiliang Tang, Philip~S Yu, and Qingsong Wen.
\newblock Large language models for education: A survey and outlook.
\newblock {\em arXiv preprint arXiv:2403.18105}, 2024.

\bibitem{wen2024onenet}
Qingsong Wen, Weiqi Chen, Liang Sun, Zhang Zhang, Liang Wang, Rong Jin, Tieniu Tan, et~al.
\newblock Onenet: Enhancing time series forecasting models under concept drift by online ensembling.
\newblock {\em Advances in Neural Information Processing Systems}, 36, 2024.

\bibitem{xu2017machine}
Jie Xu, Kyeong~Ho Moon, and Mihaela Van Der~Schaar.
\newblock A machine learning approach for tracking and predicting student performance in degree programs.
\newblock {\em IEEE Journal of Selected Topics in Signal Processing}, 11(5):742--753, 2017.

\bibitem{xu2024ai}
Tianlong Xu, Yi-Fan Zhang, Zhendong Chu, Shen Wang, and Qingsong Wen.
\newblock Ai-driven virtual teacher for enhanced educational efficiency: Leveraging large pretrain models for autonomous error analysis and correction.
\newblock {\em arXiv preprint arXiv:2409.09403}, 2024.

\bibitem{yan2024errorradar}
Yibo Yan, Shen Wang, Jiahao Huo, Hang Li, Boyan Li, Jiamin Su, Xiong Gao, Yi-Fan Zhang, Tianlong Xu, Zhendong Chu, et~al.
\newblock Errorradar: Benchmarking complex mathematical reasoning of multimodal large language models via error detection.
\newblock {\em arXiv preprint arXiv:2410.04509}, 2024.

\bibitem{yin2023survey}
Shukang Yin, Chaoyou Fu, Sirui Zhao, Ke~Li, Xing Sun, Tong Xu, and Enhong Chen.
\newblock A survey on multimodal large language models.
\newblock {\em arXiv preprint arXiv:2306.13549}, 2023.

\bibitem{mmtbench}
Kaining Ying, Fanqing Meng, Jin Wang, Zhiqian Li, Han Lin, Yue Yang, Hao Zhang, Wenbo Zhang, Yuqi Lin, Shuo Liu, Jiayi Lei, Quanfeng Lu, Runjian Chen, Peng Xu, Renrui Zhang, Haozhe Zhang, Peng Gao, Yali Wang, Yu~Qiao, Ping Luo, Kaipeng Zhang, and Wenqi Shao.
\newblock Mmt-bench: A comprehensive multimodal benchmark for evaluating large vision-language models towards multitask agi, 2024.

\bibitem{yu2024mm}
Weihao Yu, Zhengyuan Yang, Linjie Li, Jianfeng Wang, Kevin Lin, Zicheng Liu, Xinchao Wang, and Lijuan Wang.
\newblock Mm-vet: Evaluating large multimodal models for integrated capabilities.
\newblock In {\em ICML}, 2024.

\bibitem{zhang2024logora}
Huanyu Zhang, Yi-Fan Zhang, Zhang Zhang, Qingsong Wen, and Liang Wang.
\newblock Logora: Local-global representation alignment for robust time series classification.
\newblock {\em IEEE Transactions on Knowledge and Data Engineering}, 2024.

\bibitem{zhang2024beyond}
Yi-Fan Zhang, Qingsong Wen, Chaoyou Fu, Xue Wang, Zhang Zhang, Liang Wang, and Rong Jin.
\newblock Beyond llava-hd: Diving into high-resolution large multimodal models.
\newblock {\em arXiv preprint arXiv:2406.08487}, 2024.

\bibitem{zhang2024debiasing}
Yi-Fan Zhang, Weichen Yu, Qingsong Wen, Xue Wang, Zhang Zhang, Liang Wang, Rong Jin, and Tieniu Tan.
\newblock Debiasing large visual language models.
\newblock {\em arXiv preprint arXiv:2403.05262}, 2024.

\bibitem{zhang2024mme}
Yi-Fan Zhang, Huanyu Zhang, Haochen Tian, Chaoyou Fu, Shuangqing Zhang, Junfei Wu, Feng Li, Kun Wang, Qingsong Wen, Zhang Zhang, et~al.
\newblock Mme-realworld: Could your multimodal llm challenge high-resolution real-world scenarios that are difficult for humans?
\newblock {\em arXiv preprint arXiv:2408.13257}, 2024.

\bibitem{zhou2023solving}
Aojun Zhou, Ke~Wang, Zimu Lu, Weikang Shi, Sichun Luo, Zipeng Qin, Shaoqing Lu, Anya Jia, Linqi Song, Mingjie Zhan, et~al.
\newblock Solving challenging math word problems using gpt-4 code interpreter with code-based self-verification.
\newblock In {\em The Twelfth International Conference on Learning Representations (ICLR)}, 2024.

\end{thebibliography}
\bibliographystyle{plain}

\clearpage
\appendix

\section{Related Work}
\textbf{MLLMs.} 
This field has undergone significant evolution~\cite{yin2023survey, fu2023challenger,zhang2024debiasing,fu2024mme}, initially rooted in BERT-based language decoders and later incorporating advancements in LLMs. 
MLLMs exhibit enhanced capabilities and performance, particularly through end-to-end training techniques, by leveraging advanced LLMs such as GPTs~\cite{gpt4,brown2020language},
LLaMA~\cite{touvron2023llama,touvron2023llama2},  and Vicuna~\cite{chiang2023vicuna}. Recent model developments, including Flamingo~\cite{awadalla2023openflamingo}, BLIP-2~\cite{li2023blip}, InstructBLIP~\cite{dai2024instructblip}, LLaVA~\cite{liu2023visual}, Qwen-VL~\cite{bai2023qwen}, Slime~\cite{zhang2024beyond}, and VITA~\cite{fu2024vita}, bring unique perspectives to challenges such as scaling pre-training, enhancing instruction-following capabilities, and overcoming alignment issues. 
However, the performance of these models in the face of real educational scenarios has often not been revealed.

\textbf{Multimodal Benchmark.} 
With the development of MLLMs, a number of benchmarks have been built.
For instance, MME~\cite{fu2023mme} constructs a comprehensive evaluation benchmark that includes a total of 14 perception and cognition tasks. All QA pairs in MME are manually designed to avoid data leakage, and the binary choice format makes it easy to quantify.
MMT-Bench~\cite{mmtbench} scales up the dataset even further, including $31,325$ QA pairs from various scenarios such as autonomous driving and embodied AI. It encompasses evaluations of model capabilities such as visual recognition, localization, reasoning, and planning.
MME-RealWorld~\cite{zhang2024mme} contains over 29K question-answer pairs that cover 43 subtasks across 5 real-world scenarios and is the largest manually annotated benchmark to date. 
Additionally, other benchmarks focus on real-world usage scenarios~\cite{fu2024blink,bitton2023visit}, reasoning capabilities~\cite{yu2024mm,han2023coremm} and mathematical reasoning~\cite{lu2024mathvista} or correctness~\cite{yan2024errorradar}. 
However, there are widespread issues, such as data scale, annotation quality, and task difficulty, in these benchmarks, making it hard to assess the challenges that MLLMs face in the real world.

\begin{table}[htbp]
\centering
\renewcommand{\arraystretch}{1.5} 
\caption{Draft Evaluation Criteria and Scoring Standards}\label{tab:draft_cri}
\label{tab:grading-criteria}
\begin{tabularx}{\textwidth}{|p{0.2\textwidth}|X|p{0.05\textwidth}|}
\hline
\textbf{Criteria} & \textbf{Description and Scoring Standards} & \textbf{Weight (\%)} \\ \hline

\textbf{Completeness of Steps} & 
30 points: Complete solution process, including all intermediate steps.  
20-29 points: Most steps present, minor omissions.  
10-19 points: Partial steps, significant omissions.  
0-9 points: Most steps missing, incomplete process. 
& 30 \\ \hline

\textbf{Layout and Clarity} & 
25 points: Logical layout, clear writing, easy to understand.  
18-24 points: Fairly reasonable layout, mostly clear, minor ambiguities.  
10-17 points: Disorganized layout, unclear writing, harder to understand.  
0-9 points: Chaotic layout, illegible writing, incomprehensible.  
& 25 \\ \hline

\textbf{Correctness of Problem-Solving Approach} & 
20 points: Entirely correct solution approach, clear logic.  
15-19 points: Mostly correct, minor logical gaps.  
10-14 points: Significant errors or omissions affecting outcome.  
0-9 points: Incorrect solution approach, no logical reasoning.  
& 20 \\ \hline

\textbf{Logical Consistency and Rigor} & 
15 points: Rigorous logic, well-connected steps.  
10-14 points: Mostly rigorous, minor gaps.  
5-9 points: Weak logic, poor connections between steps.  
0-4 points: Chaotic logic, no clear connections.  
& 15 \\ \hline

\textbf{Unit Annotation and Answer Presentation} & 
5 points: Correct unit annotations, clear final answer.  
4 points: Mostly correct units, minor omissions.  
2-3 points: Incomplete/unclear units, vague final answer.  
0-1 points: Missing/incorrect units, unclear answer.  
& 5 \\ \hline

\textbf{Calculation Accuracy} & 
5 points: All calculations accurate.  
4 points: Most calculations accurate, minor errors.  
2-3 points: Frequent calculation errors.  
0-1 points: Severe errors, incorrect result.  
& 5 \\ \hline

\textbf{Final Score and Feedback} & 
Final score based on weighted criteria, with constructive feedback for improvement. 
& -- \\ \hline

\end{tabularx}
\end{table}

\begin{table}[htbp]
\caption{Categories and Subcategories of Student Errors with Definitions. This table presents a breakdown of the main error categories, their subcategories, and corresponding definitions, highlighting the various challenges students face during problem-solving.}
\label{tab:defi}
\centering
\resizebox{\textwidth}{!}{%
\begin{tabular}{|p{3cm}|p{3cm}|p{8cm}|}
\hline
\rowcolor{gray!20}
\textbf{Category} & \textbf{Subcategory} & \textbf{Definition} \\ \hline
\multirow{1}{4cm}{\textbf{Attitude Issues}} & Messy Drafts & Students' drafts show signs of careless scribbles. \\ \midrule
\multirow{3}{4cm}{\textbf{Misunderstanding}} & Ambiguous Statements & Problems have unclear or ambiguous wording. \\ \cline{2-3}
 & Ignoring Constraints & Students fail to notice constraints in the problem. \\ \cline{2-3}
 & Missing Key Info & Students overlook critical information in the problem. \\ \hline
\multirow{1}{4cm}{\textbf{Logical Reasoning}} & Faulty Reasoning & Students make incorrect conclusions or illogical deductions. \\ \hline
\multirow{1}{4cm}{\textbf{Cognitive Bias Errors}} & Misreading Info & Students misinterpret information due to non-habitual thinking. \\ 

\hline
\multirow{4}{4cm}{\textbf{Answering Technique}} & Improper Format & Students provide answers in an improper format. \\ \cline{2-3}
 & Draft Transcription & Calculations on the draft are correct, but transcription is wrong. \\ \cline{2-3}
 & Misaligned answer & The answer is correct, but the format is wrong. \\ \cline{2-3}
 & Incorrect Order & Students provide answers in the wrong order. \\ \hline
\multirow{1}{4cm}{\textbf{Handwriting Errors}} & Writing A, Thinking B & Students think of answer A but write down answer B. \\ \cline{2-3}
& {Digit Transcription} & Students calculate correctly but copy digits incorrectly. \\
\cline{2-3}
& {Miscounting} & Students make counting mistakes. \\
\cline{2-3}
& {Missing Units} & Students omit units in their answers. \\
\cline{2-3}
& {Incorrect Formula} & Students write down an incorrect formula. \\
\cline{2-3}
& {Extra/Missing Symbol} & Students add or omit symbols during problem-solving. \\
\cline{2-3}
& {Omitting Letters} & Students miss or add unnecessary letters in their answers. \\
\hline
\multirow{4}{4cm}{{\textbf{Attention to Detail}}} & Extra or Missing Zeros & Errors in handling numbers, such as adding or omitting zeros. \\  \cline{2-3}
& {Decimal Point Errors} & Mistakes in decimal point placement. \\  \cline{2-3}
& {Lack of Simplification} & Students fail to simplify fractions or expressions. \\  \cline{2-3}
& {Misplaced Parentheses} & Errors in using parentheses. \\  \cline{2-3}
& {Wrong Sign} & Incorrect sign usage during rearrangement. \\
\hline

\multirow{6}{4cm}{\textbf{Computation Errors}} & Arithmetic Errors & Miscalculations in addition, multiplication, or division. \\   \cline{2-3}
& {Conversion Errors} & Mistakes in converting calculation results into the final answer. \\   \cline{2-3}

& {Division Errors} & Incorrect handling of quotients or remainders. \\ \cline{2-3}
& {Decimal Multiplication} & Errors in aligning or processing decimals in multiplication. \\ \cline{2-3}
& {Fraction Comparison} & Incorrect simplification or comparison of fractions. \\
\cline{2-3}
& {Misapplied Models} & Failure to apply appropriate mathematical strategies or models. \\\hline
\multirow{1}{4cm}{\textbf{Knowledge Gaps}} & Concepts Not Mastered & Insufficient understanding or memory of essential subject. \\ \hline
\end{tabular}%
}

\end{table}

\end{document}